\documentclass[conference]{IEEEtran}
\IEEEoverridecommandlockouts
\usepackage{cite}
\usepackage{amsmath,amssymb,amsfonts}
\usepackage{algorithmic}
\usepackage{graphicx}
\usepackage{textcomp}
\usepackage{xcolor}
\usepackage{hyperref}
\usepackage{tabularx}
\usepackage{url}

\def\BibTeX{{\rm B\kern-.05em{\sc i\kern-.025em b}\kern-.08em
    T\kern-.1667em\lower.7ex\hbox{E}\kern-.125emX}}
\begin{document}

\title{Contextual Biasing to Improve Domain-specific Custom Vocabulary Audio Transcription without Explicit Fine-Tuning of Whisper Model \\
}

\author{\IEEEauthorblockN{Vishakha Lall}
\IEEEauthorblockA{\textit{Centre of Excellence in Maritime Safety} \\
\textit{Singapore Polytechnic}\\
Singapore \\
vishakha\_lall@sp.edu.sg}
\and
\IEEEauthorblockN{Yisi Liu}
\IEEEauthorblockA{\textit{Centre of Excellence in Maritime Safety} \\
\textit{Singapore Polytechnic}\\
Singapore \\
liu\_yisi@sp.edu.sg}
}

\maketitle

\begin{abstract}
OpenAI's Whisper Automated Speech Recognition model excels in generalizing across diverse datasets and domains. However, this broad adaptability can lead to diminished performance in tasks requiring recognition of specific vocabularies. Addressing this challenge typically involves fine-tuning the model, which demands extensive labeled audio data that is often difficult to acquire and unavailable for specific domains. In this study, we propose a method to enhance transcription accuracy without explicit fine-tuning or altering model parameters, using a relatively small training dataset. Our method leverages contextual biasing, to direct Whisper model's output towards a specific vocabulary by integrating a neural-symbolic prefix tree structure to guide the model's transcription output. To validate our approach, we conducted experiments using a validation dataset comprising maritime data collected within a simulated training environment. A comparison between the original Whisper models of varying parameter sizes and our biased model revealed a notable reduction in transcription word error rate and enhanced performance of downstream applications. Our findings suggest that this methodology holds promise for improving speech-to-text translation performance in domains characterized by limited vocabularies. 

\end{abstract}


\section{Introduction}
Attention-based transformer models \cite{vaswani2023attention}, renowned for their exceptional performance in Natural Language Processing (NLP), have been increasingly applied in the domain of Automated Speech Recognition (ASR) to effectively capture long-term dependencies within speech signals. Among these, OpenAI's Whisper model \cite{radford2022robust} stands out, designed with the ambitious goal of serving as a plug-and-play speech recognition system, eliminating the need for laborious supervised fine-tuning for individual applications. 

This versatility often leads to diminished performance in applications with constrained vocabularies. As a case study, the maritime communication sector, characterized by its global standardization through the Standard Maritime Communication Phrases (SMCP) \cite{smcp}, mandates a precise lexicon to ensure unambiguous communication across diverse linguistic and cultural backgrounds. Utilizing the out-of-the-box Whisper model for transcribing maritime audio snippets reveals notable inaccuracies like incorrect detection of commonly used maritime phrases, ports and locations. 

One potential solution to enhance decoder outputs for domain-specific vocabularies is fine-tuning the model. However, this approach poses a significant challenge. Achieving comparable accuracy between the general and domain-specific datasets requires the dataset to be a substantial fraction of Whisper's original training dataset, which comprises 680,000 hours of labeled audio transcription data. For specialised domains, such data is not publicly available. Collecting such extensive amounts of high-quality labeled data is not only arduous but also necessitates precise alignment between the audio and transcription. This work focuses on enhancing decoder outputs using contextual biasing techniques that do not require extensive data collection and labelling. 

To implement contextual biasing by constraining token predictions to domain-specific vocabulary, we utilize a tree-constrained pointer generator (TCPGen) \cite{see2017point} component. Unlike traditional contextual biasing methods that focus on improving the transcription of rare words using external contextual knowledge from a biasing list, TCPGen enhances the entire decoder output. TCPGen is a neural network-based component that integrates symbolic prefix-tree \cite{TAKAGI_2017} search with a neural pointer generator. This approach operates as a separate component independent of the Whisper model, enabling the use of a smaller training dataset consisting of domain-specific audio and text labels where the tokens are known to fall within the biasing list. During each inference step, TCPGen computes a probability distribution that integrates the biasing list and decoder state with the decoder's output distribution. This is accomplished by organizing the biasing list into a symbolic prefix tree and focusing on a relevant subtree at each inference step. Additionally, TCPGen calculates a generation probability that allows for the use of the original decoder output when the probability of selecting a token from the biasing list falls below a certain threshold. This approach accommodates infrequent words, such as locations and names, that may not be included in the biasing list, ensuring comprehensive and accurate transcription.

To implement our approach, we generated a list of biasing words by scraping the SMCP. This biasing list was used to train the TCPGen component without altering the model parameters on a training dataset collected in a simulated environment with participants following the communication protocols described in SMCP. We evaluated the performance by comparing the results of the transcription generated by the original and biased models both quantitatively and qualitatively. For quantitative assessment, we utilized Word Error Rate (WER), the standard metric for evaluating speech recognition performance. We observed a significant reduction in WER for the biased model across all model sizes. For qualitative validation, we compared the generated transcripts against the ground truth, and examined the impacts of the transcription errors on downstream applications. One such application involves extracting the entity the speaker is addressing and classifying whether the communication is internal to the vessel or external. Another application involves using the generated transcript to extract keywords and their corresponding timestamps to estimate the speaker's response time to environmental triggers. Our observations indicate that the inaccuracies in these applications are directly influenced by the initial transcription accuracy. In most domains, outputs of ASR serve as inputs to other analysis, therefore, the overall performance of these applications offer valuable insights into the benefits of improving the performance of the speech recognition model.

The rest of the paper is organised as follows: Sec. \ref{background} reviews the Whisper model, contextual biasing and related work. Sec. \ref{implementation} describes the implementation of the TCPGen based  biasing component. Sec. \ref{experiment} describes the experimentation setup and data preparation, followed by interpretation of results in sec. \ref{results}. Finally sec. \ref{conclusion} presents the conclusions. 

\section{Background}
\label{background}

\subsection{OpenAI Whisper ASR model}
Whisper \cite{radford2022robust} is a versatile speech recognition model designed for general-purpose use. It is trained on approximately 680,000 hours of diverse audio and excels in multilingual speech recognition, speech translation, and language identification. Unlike the Wav2Vec 2.0 model, which used unsupervised training, Whisper employs weakly supervised training with large-scale data and labels. This approach improves its ability to generalize across datasets, addressing Wav2Vec 2.0's limitation of reduced performance on some datasets due to learning dataset-specific patterns.

Whisper uses the popular transformer architecture with an attention-based encoder-decoder framework. In Section III, we analyze this architecture in detail as we develop our proposed methodology on top of it.

\subsection{Contextual Biasing}
Contextual biasing in speech recognition incorporates external information or cues relevant to the specific speech being transcribed, aiding the model in more accurately predicting contextually appropriate words and phrases. For end-to-end transformer-based ASR models, we reviewed the following approaches to implement contextual biasing.

\subsubsection{Providing an initial prompt}
\label{initial-prompt}
The Whisper model offers an optional parameter, \verb|initial-prompt|, during decoding to use fictitious prompts that guide model outputs by initializing the encoder layers with the prompt tokens' embeddings. However, this approach has two significant drawbacks: the prompt is limited to a maximum of 224 tokens, and the attention mechanism inherently assigns higher weights to tokens at the end of a longer prompt, resulting in unequal importance. This is problematic for our needs, as our vocabulary size exceeds the limit and we require equal importance for all vocabulary terms.

\subsubsection{Fine-tuning model}
Fine-tuning a pre-trained model is a common approach to enhance its utility for a specific domain. In our experiments with fine-tuning the Whisper models, we faced two major challenges. First, there was limited availability of labeled and clean data for the maritime context. High-quality data required precise alignment of text labels with audio timestamps, ensuring that smaller audio segments retained accurate labels without contamination from adjacent segments. Second, the \verb|WhisperTokenizer| did not include all words from the SMCP vocabulary. The details and results for fine-tuning are presented in Sec. \ref{training-details} and Sec. \ref{ft-results} espectively.

Another popular approach is decoder-only training, which does not require audio data and relies solely on biasing vocabulary. This method is effective for language models designed for autoregressive text generation. However, in encoder-decoder architectures like Whisper, the encoder is essential for converting audio inputs into a format compatible with the decoder. Fine-tuning only the decoder disrupts this synergy, as it begins to rely exclusively on previously generated tokens, ignoring the encoder's representations.

\subsubsection{Fusion Based Approaches}
Contextual biasing in speech recognition involves various techniques. One approach \cite{williams18_interspeech}, \cite{chen2018endtoend} uses a weighted finite-state transducer with shallow fusion, but its flexibility is limited by its reliance on specific activation terms. Another method \cite{le2021contextualized} applies deep fusion and attention, similar to the \verb|initial-prompt| method discussed in Sec. \ref{initial-prompt}, and faces challenges with large biasing vocabularies.

A different approach adds a contextual spelling correction model \cite{luitel2024contextual} on top of ASR systems, which is effective mainly for spelling errors rather than general transcription inaccuracies.

For Whisper models, CB-Whisper \cite{li2024multitask} enhances recognition of user-defined named entities using an open-vocabulary keyword-spotting mechanism, improving entity recall compared to baseline Whisper. However, this focuses on named entities rather than general token biasing.

Our work builds on the approach explored in \cite{sun2021treeconstrained}, which uses the TCPGen component for rare word biasing in ASR models. While TCPGen is designed to address infrequent entities, our approach advances this concept by applying biasing to all words, not just rare ones. This broader focus on general token biasing extends beyond TCPGen’s scope and provides a more comprehensive solution for improving ASR performance.

\section{Implementation of Contextual Biasing}
\label{implementation}

\subsection{Tree Constrained Pointer Generator}
TCPGen is a neural network component integrated with the pre-trained Whisper model, using a Graph Convolutional Neural Network (GCN). It combines a symbolic prefix-tree search with a neural pointer generator, where the prefix tree is built from a predefined biasing list. TCPGen calculates a token distribution constrained by this prefix-tree. A threshold value determines reliance on the biasing list: if the probability across all valid tokens is too low, the distribution defaults to the original model. Otherwise, the original model’s distribution is used when the generation probability is below this threshold. 

\subsubsection{Generation of Prefix-tree}
\begin{figure}[t]
\centerline{\includegraphics[width=0.15\textwidth]{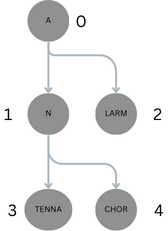}}
\caption{An example prefix tree generated for a small biasing list \{`antenna',`anchor',`alarm'\}. In memory, the structure is stored as \{'antenna':3,'anchor':4,'alarm':2\}}
\label{prefix-tree}
\end{figure}

The prefix-tree utilizes the trie data structure, which is generated from the biasing list vocabulary. Fig.~\ref{prefix-tree} illustrates a prefix-tree constructed from a sample biasing list.   

\subsubsection{Biasing of Whisper Model}
\begin{figure*}[t]
\centerline{\includegraphics[width=0.85\textwidth]{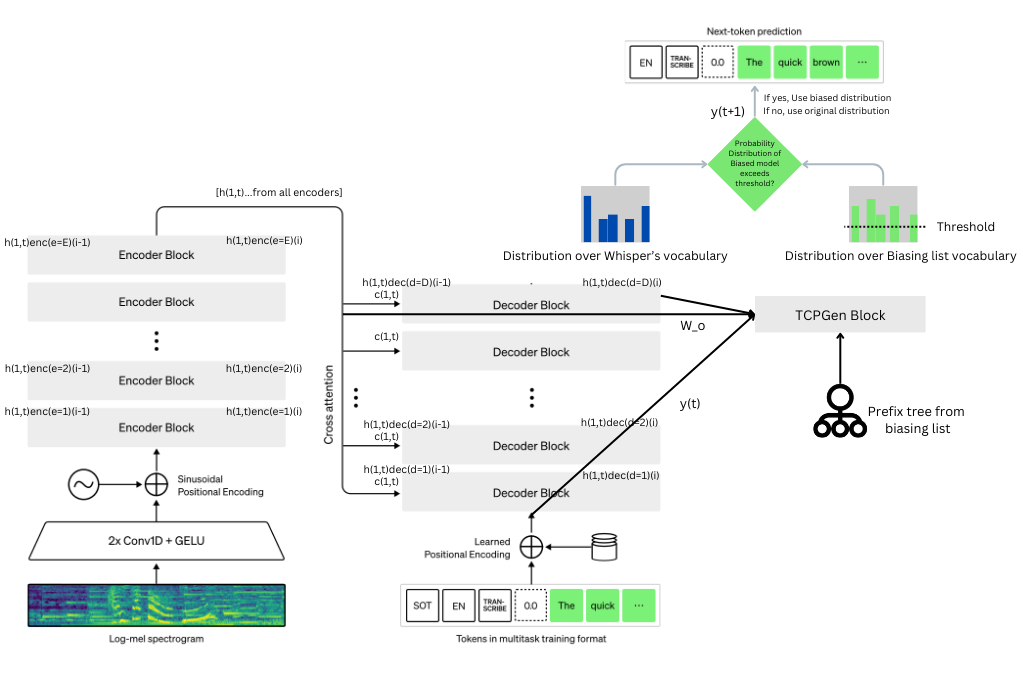}}
\caption{Whisper architecture (source: \href{https://openai.com/index/whisper/}{OpenAI Whisper Blog}) updated with contextual biasing algorithm illustrating biasing of all tokens to biasing vocabulary}
\label{whisper-architecture-with-biasing}
\end{figure*}

Fig.~\ref{whisper-architecture-with-biasing} illustrates the biasing methodology as a module connected to the Whisper model. The encoder takes a log-Mel spectrogram as input at time step $t$, and its last hidden states are fed to the decoder via cross-attention mechanisms. The decoder autoregressively predicts text tokens, conditioned on the encoder hidden states and previously predicted tokens. Each encoder layer $e={1...E}$, consisting of a self-attention block and feed-forward neural network, encodes the input and previous state $\mathbf{h_{1,t}^{enc_e(i-1)}}$ into high-level features $\mathbf{h_{1,t}^{enc_e(i)}}$. The decoder is a language model, generating text transcriptions from hidden states. At each decoder layer $d={1...D}$, a context vector $\mathbf{c_i}$ is formed using attention on $\mathbf{h_{1,t}^{enc_e(i)}}$ and the previous decoder hidden state $\mathbf{h_{1,t}^{dec_d(i-1)}}$. This context vector helps compute features $\mathbf{h_{1,t}^{dec_d(i)}}$ using previous state $\mathbf{h_{1,t}^{dec_d(i-1)}}$. The final decoder layer output passes through a Softmax layer to produce the probability distribution $P(y_{t+1}|\mathbf{y_{1,t}},\mathbf{x_{1,t}})=Softmax(\mathbf{W_o}[\mathbf{h_{1,t}^{dec_{d=D}(i)}};\mathbf{c_i}])$, where $y_{t+1}$ is the ${t+1}^{th}$ transcribed token.

At each output step $t+1$, TCPGen computes the list of valid tokens $V_{valid}$ using the generated prefix tree by searching on prefix $y_t$ (here $y_t$ is the subword generated for the previous timestamp). An \verb|exclude(v)| method sets the output to $0$ when $v \not\in V_{valid}$. The output probability distribution is then an extension of the output probability distribution generated by the final layer of the model as $P(y_{t+1}|\mathbf{y_{1,t}},\mathbf{x_{1,t}})=Softmax(\mathbf{W_o}[\mathbf{h_{1,t}^{dec_{d=D}(i)}};\mathbf{c_i}])$

To leverage the probability distribution generated over the biasing list by TCPGen, we set a threshold. A generation probability is used to interpolate between the TCPGen-generated distribution and the model-generated distribution. To prioritize the biased probability distribution and harness the patterns learned by the model, we apply a thresholding function such that if the biased probability distribution falls below the threshold, indicating that no words from the biasing list are a good match for the current output step, the output is selected from the original model's probability distribution. 

\section{Experimental Setup}
\label{experiment}

\subsection{Audio Pre-processing}
We convert audio into relevant features following the process used in the original Whisper model. To prevent hallucinations, a Voice Activity Detector (VAD) removes silences from the audio. The remaining audio and transcripts are split into 10-second segments. Audio is sampled at 16 kHz. The \verb|WhisperFeatureExtractor| converts the audio into log-Mel spectrograms, representing the signal's frequencies.

\subsection{Biasing List}
The biasing list is created by parsing the SMCP documents, extracting relevant words, and performing text cleaning to remove unwanted characters. It also includes a comprehensive list of port names, locations, harbors, and port controls extracted from sea charts.

\subsection{Training and Validation Dataset}
The dataset was collected at the \textit{Centre of Excellence in Maritime Safety, Singapore}, during simulated training exercises conducted by maritime officers adhering to the communication guidelines prescribed in the SMCP. Approximately 120 hours of audio data (including silences) were manually transcribed into text, creating the training, validation and test dataset containing 17k, 4k and 4k samples respectively.  

\subsection{Model and Training Details}
\label{training-details}
The experiments include studying the results of original, fine-tuned and contextually biased model outputs for whisper-tiny, whisper-base, whisper-small and whisper-medium with 39, 74, 244, 769 million parameters respectively. 
Fine-tuning is done using the cross-entropy loss function with different learning rates between 0.005 to 0.01 and 10000 iterations. 
The training of the TCPGen GCN component is performed independently of the Whisper models by freezing the original Whisper models' parameters at their pre-trained values. The TCPGen component was trained over 30 epochs using the Adam optimizer to optimize a Negative Loss Likelihood (NLL) loss with an adaptive learning rate starting at 0.005 and a decay rate of 20\%. The batch size for both training is 16. Training was conducted on an Nvidia RTX A5000 GPU. 

\subsection{Evaluation Metrics}
Word Error Rate (WER) is a standard metric for evaluating speech recognition systems. It is calculated as the sum of substitutions (S), deletions (D), and insertions (I) divided by the total number of words in the reference transcript (N): $WER = \frac{S+D+I}{N}$. To thoroughly assess the impact of enhanced transcription accuracy, we examine the following downstream applications.

\subsubsection{Named Entity Recognition to Extract Communication Entity, Communication Label and Locations}
One specific application of generated audio transcripts is accurately identifying entities, such as conversation addressees, internal versus external vessel entities, and mentioned locations. Transcription inaccuracies can lead to incorrect entity classification or missed detection. We use a fine-tuned BERT model for named entity recognition, supplemented by fuzzy matching to handle minor spelling variations and record detection accuracy across different experiments.

\subsubsection{Keyword Extraction and Response Time Measurement}
Another application of the generated transcript is analyzing the speaker's response time to environmental triggers simulated during training exercises. We extract keywords and their timestamps to calculate the overall response time. Transcription errors can lead to the omission of crucial keywords and inaccurate response time estimates.

\section{Results and Discussion}
\label{results}

\subsection{Fine-tuning Results}
\label{ft-results}
\begin{figure}[b]
\centerline{\includegraphics[width=0.5\textwidth]{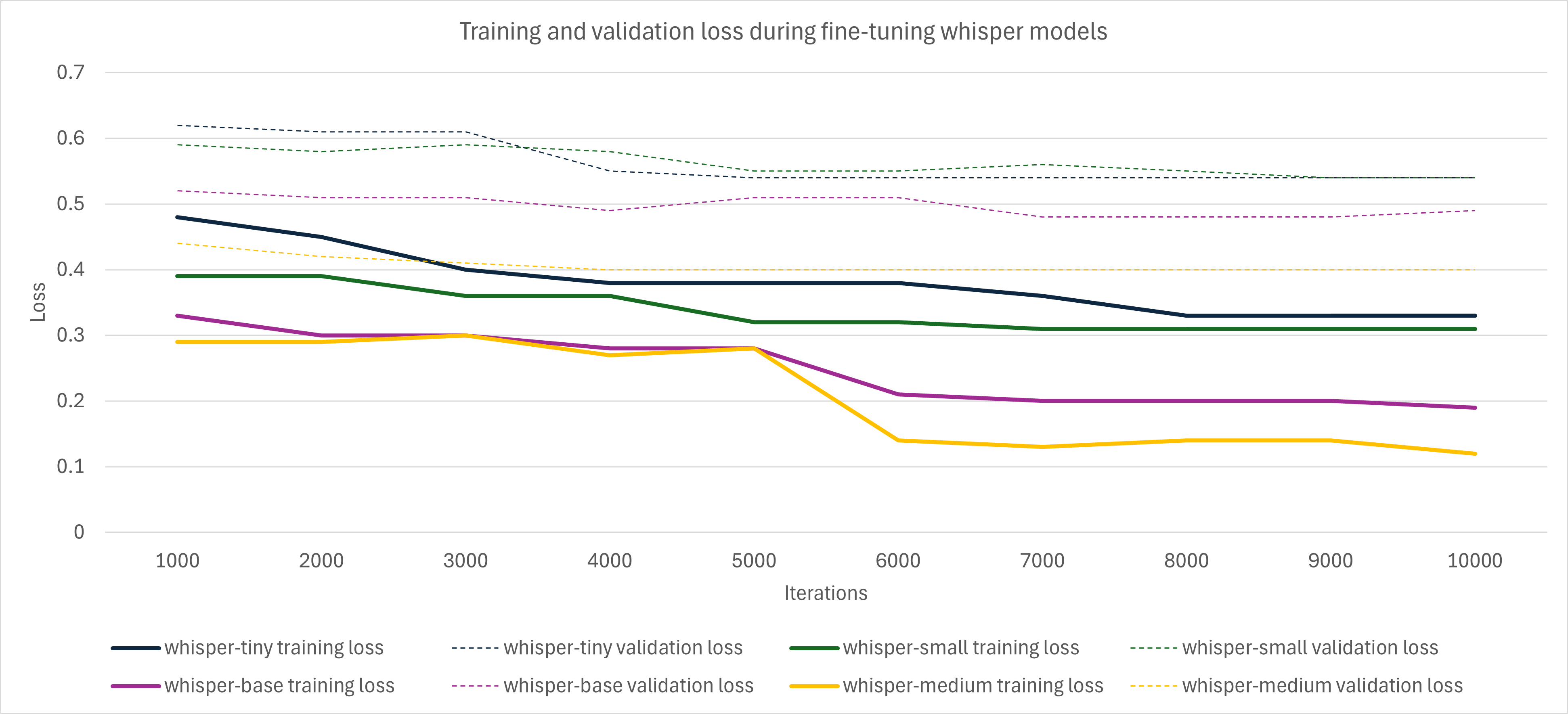}}
\caption{Plot demonstrating training and validation loss during fine-tuning of Whisper models}
\label{fine-tuning-loss}
\end{figure}

The training and validation losses from the fine-tuning experiments on all Whisper models are shown in Fig.~\ref{fine-tuning-loss}. As expected, models with more parameters have lower losses in both training and validation. While the training loss decreases significantly after about 5,000 iterations, the validation loss does not show a substantial reduction, indicating poor generalization. The overfitting is predictable because the smaller dataset is insufficient to train the encoder and decoder states. We experimented with different learning rates but observed minimal improvements, so we selected the best-performing model for further evaluation.

\subsection{TCPGen Training Results}
\begin{figure}[t]
\centerline{\includegraphics[width=0.5\textwidth]{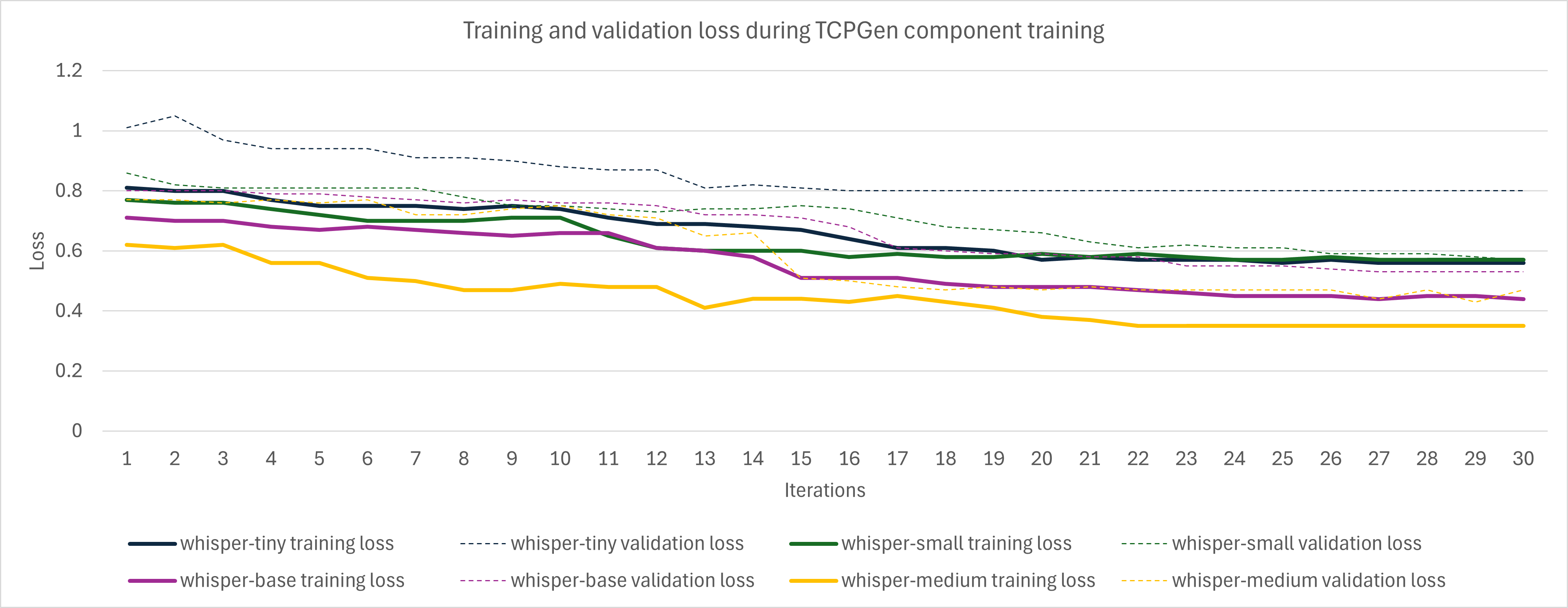}}
\caption{Plot demonstrating training and validation loss during training of TCPGen component on Whisper models}
\label{contextual-biasing-loss}
\end{figure}

Figure \ref{contextual-biasing-loss} shows the training and validation losses from the TCPGen component for all Whisper models, both of which consistently decrease with iterations. Our experiments demonstrate the effectiveness of the comprehensive biasing list, with a fallback rate to the original model outputs of only 6\%, suggesting that a smaller dataset is indeed sufficient to capture the specialized domain's linguistics.

\subsection{Quantitative Comparison on Transcription Accuracy}
The contextually biased models were evaluated against the original and fine-tuned models using WER on the test dataset. The fine-tuned models performed worse than the original models due to overfitting, while our contextual biasing implementation significantly reduced WER across all models. The comparative results are shown in Table \ref{wer-comparison}. Notably, the contextually biased Whisper-tiny and Whisper-small models perform comparably to the original Whisper-medium model. With fewer parameters, they are computationally inexpensive and faster to run, reducing processing time when using specialized vocabulary without significantly impacting performance.

\begin{table}[t]
\caption{WER comparison between original model, fine-tuned model and contextually boased model}
\begin{tabularx}{0.5\textwidth}{ 
  | >{\raggedright\arraybackslash}X 
  | >{\raggedright\arraybackslash}X
  | >{\raggedright\arraybackslash}X
  | >{\raggedright\arraybackslash}X | }
\hline
                        & \textbf{Original Whisper output} & \textbf{Fine-tuned Whisper output} & \textbf{Whisper output with TCPGen contextual biasing} \\ \hline
\textbf{whisper-tiny}   & 40.27                            & 41.89                              & 29.26                                                  \\ \hline
\textbf{whisper-small}  & 39.81                            & 41.21                              & 28.15                                                  \\ \hline
\textbf{whisper-base}   & 31.11                            & 34.23                              & 19.45                                                  \\ \hline
\textbf{whisper-medium} & 27.82                            & 31.37                              & 11.12                                                  \\ \hline
\end{tabularx}
\label{wer-comparison}
\end{table}

\subsection{Qualitative Results on Transcription Accuracy}
Table ~\ref{transcription-results} presents sample transcriptions on the test dataset produced by original whisper-medium model and contextually biased whisper-medium model against the ground truth. Upon closer examination, the qualitative benefits of contextual biasing become apparent. While the non-biased model exhibits erroneous transcriptions for locations such as `Keppel' and `Brani', the biased model accurately identifies these locations. Furthermore, the biased model demonstrates proficiency in recognizing maritime-specific terms like `starboard side' and `berthing'.

\begin{table}[t]
    \caption{Sample transcription results from test dataset}
    \begin{tabularx}{0.5\textwidth}{ 
  | >{\raggedright\arraybackslash}X 
  | >{\raggedright\arraybackslash}X 
  | >{\raggedright\arraybackslash}X | }
    \hline
        \textbf{Ground Truth} & \textbf{Transcriptions generated from original whisper-medium model} & \textbf{Transcriptions generated from TCPGen Contextual Biasing whisper-medium model} \\ 
        \hline
        Keppel Control Keppel Control this is SMA Voyager, we are headed for Brani 7 and we have a vessel crossing ahead of us. Can you give us the name of that vessel over? & \textbf{Capital control, capital control}, this is SMA Voyager. We are headed for Brani's 7 and we have a vessel crossing a \textbf{herbivast}. Can you give us the name of that vessel over? & \textbf{Keppel Control, Keppel Control}, this Is Sma Voyager. we are headed for Brani 7 and We have a vessel crossing \textbf{ahead of us}. Can you give us the name of that vessel over? \\ \hline
        Can you advise which berth is the vessel on my starboard side going to? Is it also berthing at brani over? & can you advise which \textbf{birth} is the vessel on my \textbf{star but side} going to is it also \textbf{birding} at \textbf{brownie} over? & Can You Advise which \textbf{berth} Is the Vessel On My \textbf{Starboard Side} Going to? Is It Also \textbf{berthing} At Brani Over? \\ \hline
    \end{tabularx}
    \label{transcription-results}
\end{table}

\begin{table*}[t]
    \caption{Sample entity extraction results from test dataset}
    \begin{tabularx}{\textwidth}{ 
  | >{\raggedright\arraybackslash}X 
  | >{\raggedright\arraybackslash}X
  | >{\raggedright\arraybackslash}X
  | >{\raggedright\arraybackslash}X 
  | >{\raggedright\arraybackslash}X | }
    \hline
        \textbf{Ground truth transcript} & \textbf{Transcriptions generated from original whisper-medium model} & \textbf{Extracted Entities}  & \textbf{Transcriptions generated from TCPGen Contextual Biasing whiser-medium model} & \textbf{Extracted Entities}  \\ \hline
        Copied sir pilot will be boarding on arrival. Can I know the pilot boarding arrangements sir please? & Copy sir. \textbf{Piled up} will be boarding On arrival. Can I know pulled up arrangements sir please? & - & Copy,  sir. Pilot will be boarding On arrival. Can I know the pilot boarding arrangement,  sir,  please? & Pilot \\ \hline
        VTS East, this is motor vessel adventurer. Please go ahead. & \textbf{Vide A is east}, This is Motor Vessel Adventurer. Please go ahead & Motor Vessel Adventurer & VTS east, This is Motor Vessel Adventurer. Please go ahead & VTS East, Motor Vessel Adventurer \\ \hline
    \end{tabularx}
    \label{entity-extraction-results}
\end{table*}

\begin{table}[t]
    \caption{Comparative performance metrics}
    \begin{tabularx}{0.5\textwidth}{ 
  | >{\raggedright\arraybackslash}X 
  | >{\raggedright\arraybackslash}X 
  | >{\raggedright\arraybackslash}X | }
    \hline
        \textbf{Performance metric} & \textbf{whisper-medium} & \textbf{TCPGen Contextual Biasing on whisper-medium} \\ \hline
        Word Error Rate (WER) of generated transcriptions  & 27.82 & 9.51 \\ \hline
        Classification accuracy of communication entity and communication level detections & 67\% & 98\% \\ \hline
        Response time Mean Square Error (MSE) & 0.49 & 0.11 \\ \hline
    \end{tabularx}
    \label{performance-metrics}
\end{table}

\subsection{Quantitative Results on Downstream Application Performance}
Table \ref{entity-extraction-results} illustrates the impact of transcription accuracy on downstream applications reliant on entity and keyword extraction as their initial step. It highlights how the omission of critical tokens in transcription, while potentially having minimal effect on the overall WER if the rest of the transcript is accurate, can lead to missed detections during entity extraction. For instance, in communication entity and level detection, missing tokens could obscure the addressee of the conversation. Similarly, in response time estimation, inaccurate estimates could misrepresent actual performance.

Finally, Table \ref{performance-metrics} presents a compilation of performance metrics, including the WER of transcriptions, the accuracy of communication entity and level extraction, and the mean squared error of response time estimation for whisper-medium model and the corresponding contextually biased model. Across all three metrics, we observe improvements when utilizing the TCPGen biasing.

\section{Conclusion}
\label{conclusion}

In conclusion, our study demonstrates the efficacy of contextual biasing in improving the Whisper model's performance for domain-specific applications, particularly in maritime communication. By using the TCPGen-based biasing component, we effectively constrained token predictions to align with the SMCP, reducing inaccuracies in the out-of-the-box Whisper models. Our approach, which requires neither model parameter changes nor extensive datasets, significantly reduced the WER and enhanced downstream application performance.

Our findings highlight the potential of the proposed contextual biasing approach for domains where specialized vocabulary is critical and labeled datasets are limited. Future research can explore extending our approach to similar domains.

Overall, our study advances the field of ASR, laying the foundation for more accurate and reliable speech recognition systems tailored to specific applications. This research opens avenues for extending contextual biasing techniques to enhance model performance across various specialized domains and other transformer-based speech models.

\end{document}